\documentclass[journal]{IEEEtran}
\usepackage{amsmath,amsfonts}
\usepackage{algorithmic}
\usepackage{algorithm}
\usepackage{array}
\usepackage{textcomp}
\usepackage{stfloats}
\usepackage{url}
\usepackage{verbatim}
\usepackage{graphicx}
\usepackage{cite}

\usepackage{booktabs}
\usepackage{multirow}
\usepackage[table,xcdraw]{xcolor}
\usepackage{bm}
\usepackage{graphicx}

\begin{document}

\title{BM-CL: Bias Mitigation through the lens of Continual Learning}

\author{Lucas Mansilla, Rodrigo Echeveste, Camila Gonzalez, Diego H. Milone, Enzo Ferrante

\thanks{Lucas Mansilla, Rodrigo Echeveste and Diego H. Milone are with the Research Institute for Signals, Systems and Computational Intelligence, sinc(i), CONICET - Universidad Nacional del Litoral.}
\thanks{Camila Gonzalez is with the AI Development and Evaluation (AIDE) Lab, Stanford University.}
\thanks{Enzo Ferrante is with the Institute of Computer Sciences (ICC), CONICET - Universidad de Buenos Aires.
}
\thanks{Correspondence to lmansilla@sinc.unl.edu.ar or eferrante@dc.uba.ar
}

}

\maketitle

\begin{abstract}
Biases in machine learning pose significant challenges, particularly when models amplify disparities that affect disadvantaged groups. Traditional bias mitigation techniques often lead to a {\itshape leveling-down effect}, whereby improving outcomes of disadvantaged groups comes at the expense of reduced performance for advantaged groups. This study introduces Bias Mitigation through Continual Learning (BM-CL), a novel framework that leverages the principles of continual learning to address this trade-off. We postulate that mitigating bias is conceptually similar to domain-incremental continual learning, where the model must adjust to changing fairness conditions, improving outcomes for disadvantaged groups without forgetting the knowledge that benefits advantaged groups. Drawing inspiration from techniques such as Learning without Forgetting and Elastic Weight Consolidation, we reinterpret bias mitigation as a continual learning problem. This perspective allows models to incrementally balance fairness objectives, enhancing outcomes for disadvantaged groups while preserving performance for advantaged groups. Experiments on synthetic and real-world image datasets, characterized by diverse sources of bias, demonstrate that the proposed framework mitigates biases while minimizing the loss of original knowledge. Our approach bridges the fields of fairness and continual learning, offering a promising pathway for developing machine learning systems that are both equitable and effective. 
\end{abstract}


\section{Introduction}
\label{sec:introduction}

\IEEEPARstart{M}{achine} learning systems have achieved remarkable success in a variety of tasks, ranging from automatic translation to facial recognition. However, as these technologies are increasingly deployed in society, concerns about bias and discrimination have emerged \cite{buolamwini2018gender,mehrabi2021survey}. Biases in machine learning often manifest as performance disparities between demographic groups, undermining the reliability and fairness of these systems, especially in sensitive domains such as healthcare, finance, and public policy.

Performance disparities across subgroups can arise from different sources, including data imbalance, label noise, spurious correlations, or even inherent characteristics associated with demographic groups \cite{zong2022medfair}. Data imbalance is one of the most common sources of bias \cite{larrazabal2020gender}, as many datasets lack demographic diversity. Spurious correlations arise when models exploit irrelevant features as predictive signals \cite{izmailov2022feature}. For instance, a model may learn to classify an image by leveraging background, texture, or demographic information, rather than focusing on the relevant features \cite{ye2024spurious}. Existing bias mitigation techniques, while effective in tackling group disparities, tend to suffer from the \emph{leveling-down effect}, whereby performance improvements for disadvantaged groups negatively impact the performance of advantaged groups \cite{zietlow2022leveling,mittelstadt2023unfairness}. This trade-off highlights the need for innovative approaches that promote fairness without compromising overall performance.

Continual learning, a paradigm that enables models to learn tasks sequentially without forgetting previously acquired knowledge \cite{chen2018lifelong}, offers a promising direction to address these challenges. We hypothesize that the leveling-down effect can be interpreted as a form of \emph{catastrophic forgetting} \cite{french1999catastrophic}, where optimizing for disadvantaged groups leads to a loss of knowledge about advantaged groups. To tackle this issue, we introduce \emph{Bias Mitigation through Continual Learning} (BM-CL), a bias mitigation strategy inspired by continual learning techniques such as Elastic Weight Consolidation (EWC) \cite{kirkpatrick2017overcoming} and 
Learning without Forgetting (LwF) \cite{li2017learning}. BM-CL follows a two-step training process specifically designed to correct a biased model against disadvantaged groups without compromising outcomes for advantaged groups or affecting overall performance. 

We validate our approach on both synthetic and real-world datasets widely used in bias mitigation research, demonstrating its effectiveness across diverse scenarios. Compared to baseline bias mitigation techniques, our method consistently enhances performance in disadvantaged groups without deteriorating performance in advantaged groups. By framing bias mitigation as a continual learning problem, our work paves the way for leveraging the extensive toolkit of existing continual learning methods to address fairness concerns.

\subsection{Related Work}
\label{sec:related_work}

In the last decade, a plethora of studies have demonstrated that machine learning systems can exhibit biases against specific demographic groups, often defined by protected attributes such as gender, age, or race \cite{angwin2022machine,buolamwini2018gender,larrazabal2020gender,seyyed2021underdiagnosis}. These findings have prompted growing awareness in the research community about the need to not only enhance accuracy but also improve fairness in decision-making outcomes.

Group fairness is among the most widely used definitions of algorithmic fairness in the literature \cite{dwork2012fairness}, which aims to reduce inequity in decisions across groups defined by protected attributes. In the context of binary classification tasks, fairness techniques often strive for group parity using specific metrics. Examples include demographic parity \cite{wachter2021fairness}, which ensures equal positive outcome rates among groups, or equal opportunity \cite{hardt2016equality}, which seeks to achieve uniform false negative rates across groups. Alternatively, minimax group fairness \cite{diana2021minimax} focuses on reducing the worst-case outcomes, ensuring that the group facing the greatest disparity is treated as equitably as possible, which is particularly useful in scenarios where fairness for the most vulnerable group is a primary concern. Recent studies \cite{zietlow2022leveling,mittelstadt2023unfairness,ferrante2025open} highlight a trade-off in these approaches: many current techniques for enhancing group fairness often do so at the cost of reduced performance in advantaged groups, i.e. those whose initial outcomes already exceed the average. This phenomenon, known as \textit{leveling down}, presents significant risks for machine learning technologies, particularly in critical scenarios like healthcare \cite{ricci2022addressing}, where it is ethically imperative to ensure that fairness interventions do not compromise the quality of care for any group. Forcing fairness through leveling down may result in models that, while appearing fair by reducing group differences to nearly zero, are equally harmful to all groups \cite{mittelstadt2023unfairness,sabuncu2025ethical}. In contrast, our work aligns with the principle of \textit{positive-sum fairness}, recently proposed by Belhadj et al. \cite{belhadj2024positive}, which seeks to improve outcomes for disadvantaged groups without sacrificing performance for advantaged groups. Here, we demonstrate that this goal can be achieved by reformulating the performance decrease in advantaged groups as a forgetting problem within the context of continual learning.\\

Continual learning (CL), also known as \emph{incremental} or \emph{lifelong} learning \cite{chen2018lifelong} is a paradigm that aims to mimic the human ability to learn continuously and adapt to new situations. A major challenge in this field is mitigating \textit{catastrophic forgetting} \cite{french1999catastrophic,li2017learning}, where performance on prior tasks deteriorates when learning new ones. Our work investigates the leveling down phenomenon in bias mitigation strategies as a form of catastrophic forgetting. To address forgetting in classification tasks, two primary approaches exist in CL literature: data-based techniques and prior-based techniques \cite{de2021continual}. Data-based methods extract and transfer knowledge from a prior model to a new model trained on new data. One example is the LwF method \cite{li2017learning}, which uses the predictions of the previous model as pseudo-labels for future tasks, avoiding the need to access previous task data when incorporating new ones. Prior-based approaches, on the other hand, estimate a distribution over model weights, which serves as a prior when learning with new data. Among these, EWC \cite{kirkpatrick2017overcoming} uses the Fisher information to identify the model parameters critical for solving previous tasks. In this work, we explore both LwF and EWC approaches to address the challenges of fairness and performance trade-offs.

While CL methods have demonstrated success in adapting to new tasks over time, their application to bias mitigation remains an underexplored area. The few studies in this intersection include~\cite{churamani2022domain}, which proposes a domain-incremental continual learning approach to reduce bias in facial expression and action unit recognition. This method allows models to adapt to new domains while maintaining fairness across demographic groups, but limits its exploration to regularization-based CL approaches as well as naive rehearsal. More importantly, it does not augment these methods with strategies to improve fairness. More recently, \cite{bayasi2024biaspruner} introduced \emph{BiasPruner}, which combines continual learning with bias mitigation by pruning neurons that contribute to learning spurious correlations, thus enhancing fairness in neural networks.\\

\subsection{Contributions} In this work, we introduce BM-CL, a framework to address the challenge of creating fair models by integrating task-incremental continual learning principles with bias mitigation strategies. Specifically, we show that by combining bias mitigation with LwF, we can reuse model predictions from advantaged groups in previous training stages as pseudo-labels in subsequent training to preserve knowledge for these groups, while optimizing for disadvantaged ones. Furthermore, EWC allows constraining weights that are key to performance in advantaged groups, allowing less significant weights to adjust and improve outcomes for disadvantaged groups. Experimental results demonstrate that our approach effectively improves worst-group performance while consistently preventing performance degradation in advantaged groups, aligning with the principle of positive-sum fairness.

\section{Integrating bias mitigation and continual learning}
\label{sec:methods}

\subsection{Preliminaries}

We consider a supervised classification problem where the goal is to predict a label $y \in \mathcal{Y}$ for a given input $\mathbf{x} \in \mathcal{X}$. To achieve this, we train a model $f(\mathbf{x}; \theta) : \mathcal{X} \to \mathcal{Y}$, which is parameterized by $\theta \in \Theta$, using a dataset of $n$ samples, denoted as $\{(\mathbf{x}_i, y_i)\}_{i=1}^n$. Here, $\mathbf{x}_i \in \mathcal{X}$ represents the input features and $y_i \in \mathcal{Y}$ is the corresponding target label. 

The standard framework for training supervised learning models is empirical risk minimization (ERM) \cite{vapnik1999overview}. ERM seeks to optimize the model parameters $\theta$ by minimizing the empirical risk, which corresponds to the average loss over the training set. Although ERM is effective in optimizing overall accuracy, it does not ensure fair performance across subgroups, particularly in datasets with imbalances or spurious correlations \cite{sagawa2019distributionally}. In such scenarios, the optimization objective may be dominated by majority groups, leading to poor performance for minority groups. Furthermore, models trained under spurious correlations may rely on group-specific patterns in training data, resulting in reduced performance for groups where such correlations do not hold.

Bias and group disparities can be addressed during training through various approaches, such as group rebalancing, adversarial learning, domain independence, or even domain generalization methods \cite{zong2022medfair}. A widely adopted strategy is to prioritize maximizing worst-group performance, as minority groups often experience disproportionately higher error rates. In this framework, each instance $(\mathbf{x}_i,y_i)$ is associated with a group $g_i \in \mathcal{G}$, where $\mathcal{G} = \mathcal{A} \times \mathcal{Y}$, with $a \in \mathcal{A}$ representing an attribute of interest. 

Unlike ERM, which optimizes overall accuracy, methods like \emph{Group Distributionally Robust Optimization} (GroupDRO) \cite{sagawa2019distributionally} explicitly minimize the worst-group error, ensuring that underrepresented groups are not overlooked during training. Group rebalancing methods provide another approach to bias mitigation. For example, resampling methods \cite{idrissi2022simple}, referred to as \textit{ReSample} in this work, aim to adjust group contributions by upsampling minority groups. This prevents the model from prioritizing groups with larger representation during training. Another resampling method is Just Train Twice (JTT) \cite{liu2021just}, which follows a two-stage approach. In the first stage (\textit{identification}), a model is trained using ERM to identify samples with high error which are likely to belong to worst-performing groups. In the second stage (\textit{upweighting}), the identified samples are reweighted during training to specifically improve the performance of these groups. 

In this work, we adopt these methods as baselines to illustrate their leveling down effect and demonstrate how this issue can be mitigated by integrating CL strategies into the training process.

\subsection{Two-stage framework}

\begin{figure*}
    \centering
    \includegraphics[width=\linewidth]{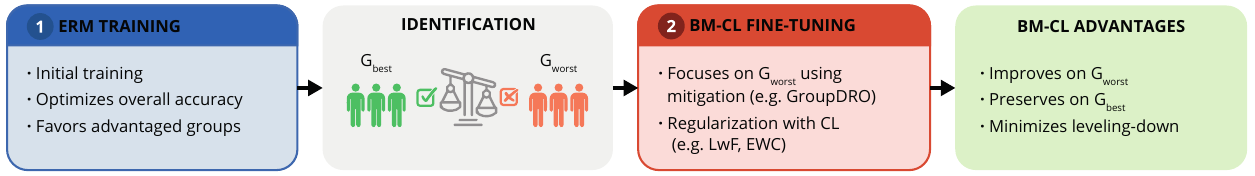}
    \caption{Overview of the proposed BM-CL framework, composed of two-stages. First, we train a model in the standard way (by ERM) and identify best and worst groups ($\mathcal{G}_{best}$ and $\mathcal{G}_{worst}$). Then, we fine-tune the model using bias mitigation to improve performance on $\mathcal{G}_{worst}$ and continual learning to preserve performance on $\mathcal{G}_{best}$.}
    \label{fig:BM-CL}
\end{figure*}

We hypothesize that the leveling down effect can be interpreted as a form of forgetting, where efforts to optimize fairness unintentionally degrade performance on advantaged groups. We then define BM-CL, a two-stage training process which combines CL with bias mitigation methods (Fig. \ref{fig:BM-CL}).

In the first stage, we employ standard ERM to train the model on the original data distribution, after which the best and worst performing groups are identified. In the second stage, we fine-tune the model on the same dataset with a targeted focus on improving performance for disadvantaged groups, while leveraging CL to avoid forgetting the advantaged ones. This idea resembles JTT in the sense that we are also introducing a two-step training process. However, whereas JTT simply upweights the influence of misclassified samples in the second stage, we propose to employ CL mechanisms to do so while avoiding the forgetting effect on the advantaged groups. A detailed explanation of each stage is provided below:\\

\subsubsection{Baseline training with ERM}

In the first stage, the model is trained to achieve high overall accuracy. The classical ERM objective is defined as

\begin{equation} 
    \mathcal{L}_{\text{ERM}}(\theta) = \frac{1}{n} \sum_{i=1}^n \ell(f(\mathbf{x}_i; \theta), y_i), \label{eq:loss_erm} 
\end{equation}

\noindent where $\ell: \mathcal{Y} \times \mathcal{Y} \to \mathbb{R}^+$ is the loss function (e.g. cross-entropy loss) that quantifies the discrepancy between the predicted output $f(\mathbf{x}_i;\theta)$ and the true label $y_i$. 

We empirically observed that ERM fits advantaged groups faster than disadvantaged groups, which aligns with similar observations made in previous studies \cite{nam2020learning,liu2021just}. Thus, we limit the ERM training to a fraction of the total number of epochs. This fraction is controlled by a hyperparameter $\rho$, referred to as the pretraining ratio.

After training, the model is evaluated on the validation set to compute the accuracy $\alpha_g$ for each group $g$. Based on this evaluation, groups are partitioned into two disjoint subsets: the best-performing groups ($\mathcal{G}_{\text{best}}$) and the worst-performing groups ($\mathcal{G}_{\text{worst}}$). That is,

\begin{equation}    
    \mathcal{G}_\text{best} = \{g \mid \alpha_g > \tau \},    \quad 
    \mathcal{G}_\text{worst} = \{g \mid \alpha_g \leq \tau \}.
\end{equation}

\noindent The partition is determined by a threshold $\tau$ representing the balanced accuracy, i.e. the mean accuracy across all groups,

\begin{equation}    
    \tau=\frac{1}{|\mathcal{G}|} \sum_{g \in \mathcal{G}} \alpha_g,
\end{equation}

\noindent with $|\mathcal{G}|$ denoting the total number of groups. This procedure classifies the groups into those that perform above and below the average accuracy threshold, providing a basis for the subsequent stage.\\

\subsubsection{Fine-tuning for bias mitigation with continuous learning}

In the second stage, we propose to fine-tune the model using a traditional bias mitigation method (e.g., GroupDRO or ReSample), but regularizing them to avoid forgetting via CL. The goal is to improve performance for disadvantaged groups ($g \in \mathcal{G}_{\text{worst}}$) while preserving knowledge of the best-performing groups ($g \in \mathcal{G}_{\text{best}}$). We do so by optimizing the objective

\begin{equation} 
    \mathcal{L}_{\text{BM-CL}}(\theta) = \mathcal{L}_{\text{BM}}(\theta) + \lambda \mathcal{L}_{\text{CL}}(\theta),
    \label{eq:loss_group_cl}
\end{equation}

\noindent where $\mathcal{L}_{\text{BM}}$ is the loss associated with the bias mitigation strategy, $\mathcal{L}_{\text{CL}}$ is the continual learning regularizer, and $\lambda$ determines the relative importance of the continual learning regularization compared to the bias mitigation loss. For $\mathcal{L}_{\text{BM}}$, we considered GroupDRO and ReSample as bias mitigation strategies; however, the framework is modular, so other methods could be considered. For $\mathcal{L}_{\text{CL}}$ we propose Learning without Forgetting (LwF) and Elastic Weight Consolidation (EWC) as the CL methods:\\

\noindent \textbf{Learning without forgetting}: LwF employs knowledge distillation \cite{hinton2015distilling} to retain the previous predictions of selected groups. Here, it is implemented using the Kullback-Leibler (KL) divergence \cite{mackay2003information} between the current and previous predictions for the samples of the best-performing groups,

\begin{equation}
    \mathcal{L}_{\text{CL}}(\theta) = \frac{1}{|\mathcal{I}_{\text{best}}|} \sum_{i \in \mathcal{I}_{\text{best}}} \text{KL}(q_i^*, q_i),
\end{equation}

\noindent where $\mathcal{I}_{\text{best}} = \{i \mid g_i \in \mathcal{G}_{\text{best}}\}$ denotes the set of indices of the best-performing samples. The predicted probabilities from the previous model ($\theta^*$) and the current model ($\theta$) are computed by applying the softmax function $\sigma$ to model outputs, 

\begin{equation}
    q_i^* = \sigma\left(\frac{f(\mathbf{x}_i; \theta^*)}{T}\right), \quad 
    q_i = \sigma\left(\frac{f(\mathbf{x}_i; \theta)}{T}\right),
\end{equation}

\noindent where $f(\mathbf{x}_i; \theta^*)$ is the output of the previous model and $f(\mathbf{x}_i; \theta)$ is the output of the current model. $T$ is the temperature parameter that smooths the probability distributions. LwF ensures that the model retains its predictive performance for the best-performing groups while adapting to the worst-performing groups. The distillation loss aligns the outputs for best-performing groups across stages, maintaining stability and reducing the risk of performance degradation.\\

\noindent \textbf{Elastic Weight Consolidation}: EWC introduces a regularization term to prevent significant changes to parameters critical for the best-performing groups:

\begin{equation}
    \mathcal{L}_{\text{CL}}(\theta) = \frac{1}{2} \sum_{j=1}^{|\theta|} F_j(\theta_j - \theta_j^{*})^2,
\end{equation}

\noindent where $\theta_j^{*}$ represents the $j$-th neural weight learned during the first stage, $\theta_j$ refers to the $j$-th weight currently being optimized, and $F_j$ corresponds to $j$-th entry in the diagonal of the Fisher information matrix \cite{jaynes2003probability}, which quantifies the importance of the corresponding weight $\theta_j$. In the context of bias mitigation, we postulate that EWC helps to balance performance between the best-performing and the worst-performing groups by selectively penalizing changes to parameters critical for best-performing groups. By doing so, it minimizes the risk of degrading accuracy on advantaged groups while allowing updates to optimize for worst-performing groups. The Fisher information is computed for samples of the best-performing groups ($\mathcal{G}_{\text{best}}$), estimating the second derivative of the logarithm of the most likely predicted label for each sample \cite{van2019three}, and serves as a measure of parameter sensitivity, ensuring that the model adapts in a way that retains prior knowledge.\\

To summarize, we formulate fairness as an incremental constraint on the distribution of decisions of the model during training. In this framework, each adjustment made to reduce bias between groups can be considered as a new task (which is how training stages are defined in continual learning literature), where the model is required to improve fairness over time while retaining previous learned capabilities. This approach aligns with the stability-plasticity trade-off of continual learning \cite{wang2024comprehensive}, allowing the use of continual learning techniques to balance performance preservation with the progressive reduction of biases.

\section{Experiments}
\label{sec:experiments}

\subsection{Datasets}

We evaluate the proposed framework on a combination of synthetic and real-world datasets, each exhibiting different sources of bias.  Below, we provide a detailed description of each dataset:

\begin{itemize}
    \item \textit{Waterbirds}: this dataset \cite{wah2011caltech, sagawa2019distributionally} consists of bird images annotated with two class labels, \emph{waterbird} and \emph{landbird}. Each image is also associated with a background attribute (\emph{water} or \emph{land}) to represent the environment where the bird is placed. The dataset is explicitly designed to simulate spurious correlations between the background (attribute) and the bird type (label), which results in a significant performance disparity when this association is broken. We use the publicly available train, validation, and test splits.
   
    \item \textit{CelebA}: the CelebFaces Attributes (CelebA) dataset \cite{liu2015deep} is a large-scale facial attribute dataset containing over $200,000$ images of celebrity faces, annotated with $40$ binary attributes such as hair color, eyeglasses, gender, and facial expressions. For our experiments, we focus on the binary classification task of predicting whether a person has blond hair. The dataset also includes demographic attributes such as gender (male or female), which introduce spurious correlations with the label. We also use the standard train, validation, and test splits for training and evaluation.
   
    \item \textit{CheXpert}: is a large-scale and multi-label dataset of chest X-ray images \cite{irvin2019chexpert}, comprising $224,316$ chest radiographs of $65,240$ patients with both frontal and lateral views annotated with $14$ medical observations. In our work, we focus on the task of detecting the presence or absence of ``Pleural effusion''. We use patient age as a demographic attribute, grouping patients into three age categories: young ($<40$), middle-aged ($\leq 65$) and old ($>65$). We use only frontal images and randomly split the dataset into $70\%$ for training, $10\%$ for validation, and $20\%$ for testing, ensuring that no patient appears in more than one split. 
\end{itemize}

\noindent These datasets were chosen to capture diverse sources of bias. Specifically, Waterbirds and CelebA are characterized by spurious correlations between attributes (e.g., background, gender) and class labels. Conversely, CheXpert highlights attribute imbalance, particularly in the representation of demographic subgroups. Table \ref{tab:datasets_summary} provides a detailed summary of key statistics for each dataset, including the number of samples in the training, validation, and test sets, as well as the sizes of the largest and smallest groups. 

\begin{table}[tb!]
\centering
\caption{Summary of datasets used in our experiments. The maximum and minimum group sizes refer to the training set.}
\begin{tabular}{lrrrrr}
\toprule
\textbf{Dataset} & \textbf{Training} & \textbf{Max group} & \textbf{Min group} & \textbf{Classes} & \textbf{Attrs} \\
\midrule
Waterbirds & 4,795 & 3,498 & 56 & 2 & 2 \\
CelebA & 162,770 & 71,629 & 1,387 & 2 & 2 \\
CheXpert & 133,785 & 35,502 & 5,521 & 2 & 6 \\
\bottomrule
\end{tabular}
\label{tab:datasets_summary}
\end{table}

\subsection{Implementation details}

\begin{table*}[t!]
\centering
\caption{Comparison of BM-CL against the ERM baseline and state-of-the-art bias mitigation methods across datasets. The best result for each metric is highlighted in bold and the smallest degradation in best-group accuracy is highlighted in blue. LDE: Leveling-down Effect; IW: Improvement Worst. Results within 0.1 of the best are treated as comparable and marked with †.}
\resizebox{\textwidth}{!}{
\begin{tabular}{lllccccccc}
\toprule
{\bf Dataset} & \multicolumn{2}{c}{\bf Method} & {\bf Global Acc.} & {\bf Balanced Acc.} & {\bf Best Group} & {\bf Worst Group} & {\bf ↓ Disparity} & {\bf ↓ LDE} & {\bf ↑ IW} \\
\midrule
\multirow{8}{*}{}
& Baseline & ERM & 88.2 $\pm$ 0.5 & 86.6 $\pm$ 0.6 & \textbf{99.5} $\pm$ 0.1 & 72.8 $\pm$ 1.3 & 26.7 & -- & -- \\
\cmidrule{2-10}
& & GroupDRO & \textbf{91.5} $\pm$ 0.3 & \textbf{90.2} $\pm$ 0.2 & 98.6 $\pm$ 0.2 & 82.6 $\pm$ 0.4 & 16.0 & 0.9 & 9.8 \\
& BM & ReSample & 90.5 $\pm$ 0.9 & 89.7 $\pm$ 0.3 & 94.9 $\pm$ 1.1 & \textbf{85.5} $\pm$ 1.5 & \textbf{9.5} & 4.6 & \textbf{12.6} \\
Waterbirds 
& & JTT & 88.8 $\pm$ 0.6 & 88.7 $\pm$ 0.3 & 96.2 $\pm$ 0.5 & 83.5 $\pm$ 1.0 & 12.8 & 3.3 & 10.7 \\
\cmidrule{2-10}
& & GroupDRO-LwF & 90.0 $\pm$ 0.6 & 89.3 $\pm$ 0.4 & 99.0 $\pm$ 0.2 & 81.6 $\pm$ 1.0 & 17.4 & 0.5 & 8.8 \\
& BM-CL & GroupDRO-EWC & 90.2 $\pm$ 0.5 & 89.4 $\pm$ 0.4 & 99.0 $\pm$ 0.3 & 81.2 $\pm$ 1.2 & 17.8 & 0.5 & 8.4 \\
& (ours) & ReSample-LwF & 89.6 $\pm$ 0.6 & 88.8 $\pm$ 0.3 & 99.3 $\pm$ 0.2 & 79.5 $\pm$ 1.5 & 19.8 & {\color{blue}\textbf{0.2}}\textsuperscript{†} & 6.7 \\
& & ReSample-EWC & 90.9 $\pm$ 0.4 & 89.2 $\pm$ 0.1 & 99.3 $\pm$ 0.2 & 78.5 $\pm$ 1.4 & 20.7 & {\color{blue}\textbf{0.2}}\textsuperscript{†} & 5.7 \\
\midrule
\multirow{8}{*}{}
& Baseline & ERM & \textbf{95.5} $\pm$ 0.1 & 82.1 $\pm$ 0.6 & \textbf{99.3} $\pm$ 0.1 & 46.1 $\pm$ 2.2 & 53.2 & -- & -- \\
\cmidrule{2-10}
& & GroupDRO & 93.5 $\pm$ 0.4 & 88.4 $\pm$ 1.0 & 95.8 $\pm$ 0.3 & 72.1 $\pm$ 3.3 & 23.7 & 3.5 & 26.0 \\
& BM & ReSample & 91.9 $\pm$ 0.3 & 89.4 $\pm$ 0.6 & 92.9 $\pm$ 0.4 & 80.1 $\pm$ 2.4 & 12.8 & 6.4 & 34.0 \\
CelebA 
& & JTT & 88.8 $\pm$ 0.4 & 86.8 $\pm$ 0.6 & 90.8 $\pm$ 1.1 & 76.7 $\pm$ 0.7 & 14.1 & 8.5 & 30.6 \\
\cmidrule{2-10}
& & GroupDRO-LwF & 93.9 $\pm$ 0.4 & 88.7 $\pm$ 0.6 & 96.5 $\pm$ 0.5 & 72.4 $\pm$ 2.4 & 24.0 & \color{blue}\textbf{2.8} & 26.3 \\
& BM-CL & GroupDRO-EWC & 93.7 $\pm$	0.2 & 89.3 $\pm$ 0.4 & 95.9 $\pm$ 0.4 & 75.2 $\pm$ 1.6 & 20.6 & 3.5 & 29.1 \\
& (ours) & ReSample-LwF & 92.5 $\pm$ 0.4 & 90.0 $\pm$ 0.3 & 94.3 $\pm$ 0.5 & 80.8 $\pm$ 1.2 & 13.5 & 5.1 & 34.7 \\
& & ReSample-EWC & 92.1 $\pm$ 0.3 & \textbf{90.2} $\pm$ 0.3 & 93.6 $\pm$ 0.3 & \textbf{82.2} $\pm$ 1.3 & \textbf{11.4} & 5.7 & \textbf{36.1} \\
\midrule
\multirow{8}{*}{}
& Baseline & ERM & 76.0 $\pm$ 0.7 & 75.5 $\pm$ 0.7 & \textbf{84.1} $\pm$ 1.6 & 67.1 $\pm$ 2.0 & 17.1 & --  & -- \\
\cmidrule{2-10}
& & GroupDRO & 76.4 $\pm$ 0.4 & 76.3 $\pm$ 0.6 & 81.3 $\pm$ 0.9 & \textbf{73.6} $\pm$ 1.1\textsuperscript{†} & \textbf{7.7} & 2.8 & \textbf{6.5}\textsuperscript{†} \\
& BM & ReSample & 76.1 $\pm$ 0.3 & 76.2 $\pm$ 0.1 & 81.0 $\pm$ 1.0 & 72.7 $\pm$ 1.3 & 8.3 & 3.1 & 5.6 \\
CheXpert & & JTT & 71.6 $\pm$ 1.2 & 71.6 $\pm$ 1.0 & 79.0 $\pm$ 1.4 & 67.3 $\pm$ 1.1 & 11.7 & 5.1 & 0.2 \\
\cmidrule{2-10}
& & GroupDRO-LwF & \textbf{77.2} $\pm$ 0.2\textsuperscript{†} & \textbf{77.1} $\pm$ 0.2\textsuperscript{†} & 83.6 $\pm$ 0.9 & 73.0 $\pm$ 1.7 & 10.6 & \color{blue}\textbf{0.5} & 5.9 \\
& BM-CL & GroupDRO-EWC & 76.2 $\pm$ 0.6 & 76.0 $\pm$ 0.5 & 81.8 $\pm$ 0.7 & 71.7 $\pm$ 0.6 & 10.2 & 2.3 & 4.6 \\
& (ours) & ReSample-LwF & \textbf{77.3} $\pm$ 0.1\textsuperscript{†} & \textbf{77.2} $\pm$ 0.3\textsuperscript{†} & 82.8 $\pm$ 1.6 & \textbf{73.5} $\pm$ 2.0\textsuperscript{†} & 9.3 & 1.3 & \textbf{6.4}\textsuperscript{†} \\
& & ReSample-EWC & 76.0 $\pm$ 0.5 & 76.0 $\pm$ 0.6 & 81.5 $\pm$ 1.4 & 72.8 $\pm$ 0.9 & 8.8 & 2.6 & 5.7 \\
\bottomrule
\end{tabular}}
\label{tab:sota_comparison_overall}
\end{table*}

For all datasets, we use a ResNet-50 architecture \cite{he2016deep} pre-trained on ImageNet \cite{russakovsky2015imagenet} as the feature extractor. The final classification layer is replaced with a fully connected layer with two output units to perform the binary classification task. All experiments are implemented in PyTorch \cite{paszke2017automatic} and executed on an NVIDIA Titan X GPU.\footnote{Code is publicly available at \url{https://github.com/lamansilla/BM-CL}}. 

We train all models to minimize the standard cross-entropy loss using stochastic gradient descent with momentum of $0.9$ and weight decay of $10^{-4}$. We fix the number of training epochs to $30$ for Waterbirds and $50$ for both CelebA and CheXpert. The validation set is used for hyperparameter tuning and model selection, considering the best worst-group accuracy as the selection criterion. We also apply early stopping with a patience of $10$ epochs to prevent overfitting. 
The batch size is set to $32$ for all experiments. The learning rate is set to $10^{-3}$ for Waterbirds and CheXpert, and $10^{-4}$ for CelebA.

\subsection{Baseline models and performance evaluation} 

We evaluate different variants of the proposed BM-CL framework incorporating LwF and EWC, and compare them to several standard baselines. These baselines include the standard ERM trained until convergence, and state-of-the-art bias mitigation methods such as GroupDRO, ReSample, and JTT. Model performance is evaluated over 5 independent runs, each initialized with a different random seed to account for variability. We report two global performance metrics: the global accuracy, which measures overall accuracy on the full test set regardless of group membership, and the aforementioned balanced accuracy, defined as the mean of group-wise accuracies, giving equal weight to each group regardless of its size. 

In addition to global metrics, we also report the accuracy of the best- and worst-performing groups, as well as the disparity between them. We further quantify the improvement and degradation in accuracy for best and worst groups, respectively, compared to the ERM baseline. The best and worst groups are determined based on the group-wise accuracies obtained from the ERM model. These group identities are kept fixed for all subsequent comparisons in order to evaluate the effect of bias mitigation strategies on the groups that were originally advantaged or disadvantaged under standard ERM training.

\section{Results}

\subsection{Comparison with baselines and state-of-the-art methods}

Table \ref{tab:sota_comparison_overall} presents a comprehensive comparison across 5 independent runs between BM-CL and state-of-the-art bias mitigation methods on the aforementioned datasets: Waterbirds, CelebA and CheXpert. We report global accuracy, balanced accuracy, group-wise accuracies (for the best- and worst-performing groups), and the disparity between them. Recall that our primary goal is to improve worst-group accuracy while ensuring that best group accuracy does not degrade compared to the ERM baseline. To better highlight this trade-off, we include two relative measures in the last two columns: the leveling-down effect, which quantifies the degradation in best-group performance relative to ERM, and the improvement in worst-group, which captures gains over ERM in worst-group performance. Overall, across all datasets, the proposed BM-CL framework (particularly with LwF regularization) consistently achieves the lowest leveling-down effect (highlighted in blue) while delivering substantial gains in worst-group accuracy. Notably, BM-CL’s improvements in worst-group accuracy tend to be on par with those of other bias-mitigation strategies, yet without causing a more pronounced leveling-down degradation.

In Waterbirds, GroupDRO achieves the highest global (91.5\%) and balanced (90.2\%) accuracies. However, ReSample offers greater consistency across groups, achieving the highest worst-group accuracy (85.5\%) and the lowest disparity. BM-CL methods perform competitively: ReSample-LwF and ReSample-EWC nearly match ERM in best-group accuracy (99.3\% vs. 99.5\%), while GroupDRO-LwF improves worst-group accuracy to 81.6\% with minimal degradation in the best group (99.0\%). Both approaches exhibit the least pronounced leveling-down effect when compared with baselines. In addition, GroupDRO-EWC also demonstrates favorable trade-offs, with comparable worst-group performance and similarly small degradation in best-group accuracy.

\begin{figure*}[t!]
    \centering
    \includegraphics[width=\linewidth]{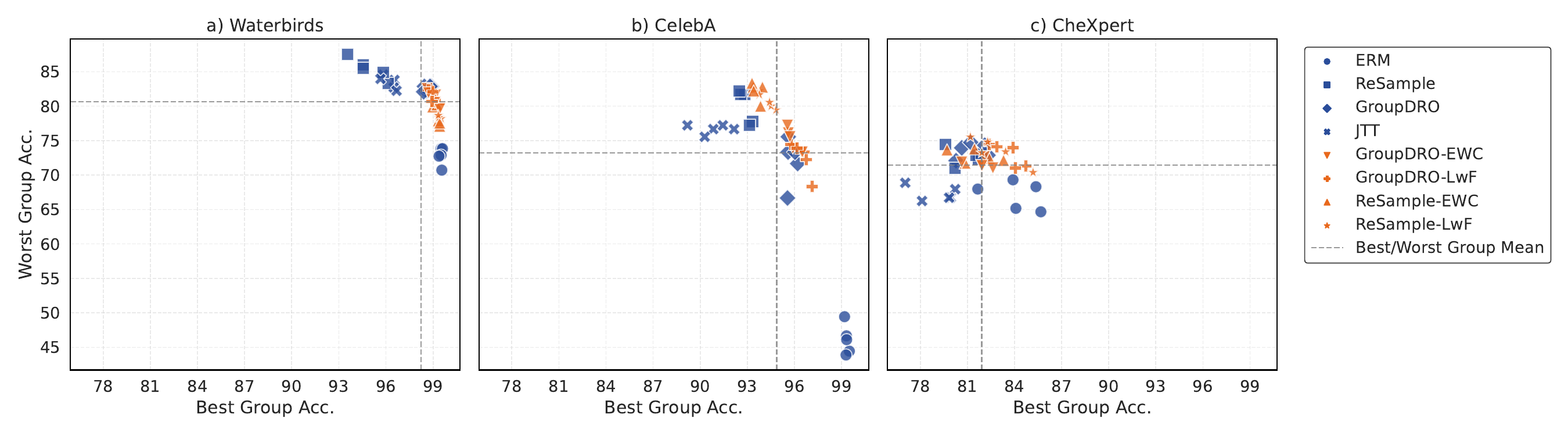}
    \caption{Comparison of the accuracy for each run and for best and worst groups across datasets. ERM and traditional bias mitigation methods (GroupDRO, ReSample, and JTT, in blue) are contrasted with BM-CL (GroupDRO-LwF and ReSample-LwF, in orange).}
    \label{fig:best_worst_comparison}
\end{figure*}

In CelebA, ERM obtains the highest global accuracy (95.5\%) and best-group accuracy (99.3\%), but suffers from a large performance gap, with worst-group accuracy dropping to 46.1\%. While methods like GroupDRO, ReSample, and JTT significantly reduce this disparity, they do so at the expense of best-group performance. In contrast, BM-CL methods offer a better balance: GroupDRO-LwF, for example, maintains high best-group accuracy (96.5\%) while improving worst-group performance, with the smallest leveling-down effect. ReSample-EWC, on the other hand, achieves the highest worst-group accuracy (82.2\%) and better preserves best-group performance (93.6\%) compared to ReSample alone (92.9\%).

In CheXpert, ERM again exhibits a considerable disparity between best- and worst-performing groups (84.1\% vs. 67.1\%). GroupDRO and ReSample both improve the worst-group accuracy (up to 73.6\% and 72.7\%, respectively), but BM-CL methods yield the most balanced results overall. GroupDRO-LwF and ReSample-LwF achieve the highest global and balanced accuracies (both around 77.0\%) and maintain the worst-group performance (73.0\% and 73.5\%). Notably, GroupDRO-LwF also yields the smallest degradation in best-group accuracy (83.6\%). GroupDRO-EWC and ReSample-EWC also perform robustly across metrics, achieving competitive worst-group performance. 

Tables \ref{tab:sota_waterbirds_groups}, \ref{tab:sota_celeba_groups}, and \ref{tab:sota_chexpert_groups} detail subgroup-level accuracies for each dataset, providing a more granular view of how each method performs across different demographic or contextual combinations in terms of fairness and performance preservation. For Waterbirds, both ReSample-LwF and GroupDRO-LwF improve accuracy in the mismatched subgroups (Wb-L and Lb-W), with minimal leveling-down on the dominant group (Lb-L). In the case of CelebA, the largest disparity appears between blond males (Bh-M) and non-blond males (Nh-M). BM-CL boosts Bh-M accuracy from 46.1\% (ERM) to 80.8\% (ReSample-LwF) and 82.2\% (ReSample-EWC), while keeping high accuracy in Nh-M (94.3\% and 93.6\%, respectively). 

In CheXpert, performance is more distributed across subgroups. BM-CL leads to significant gains in underperforming cases like young patients with pleural effusion (Pe-Y), reaching 72.4\% (GroupDRO-LwF) and 73.8\% (ReSample-LwF), compared to ERM (67.5\%). Accuracy in the best-performing group, young patients without pleural effusion (Ne-Y), remains high for BM-CL, with minimal leveling-down. GroupDRO-EWC and ReSample-EWC also improve Pe-Y performance (71.6\% and 72.0\%, respectively), supporting the utility of BM-CL in complex medical datasets.

In addition, Fig. \ref{fig:best_worst_comparison} provides a comparative view of best- and worst-group accuracy across multiple runs and datasets. Each point in the scatter plot represents a single run, with the x-axis showing the accuracy on the best-performing group and the y-axis showing the accuracy on the worst-performing group. We use color to distinguish methods that incorporate continual learning (CL) from those that do not, and include dashed lines to indicate the mean accuracy along each axis for visual reference. The figures show the trade-off between mitigating bias and avoiding the leveling-down effect. Traditional methods often succeed in boosting underperforming groups but struggle to preserve accuracy on groups that already perform well. In contrast, BM-CL variants achieve a more favorable trade-off. This is evidenced by the fact that points corresponding to CL methods mostly stay at the upper-right part of the plot, indicating that they not only mitigate bias effectively but also minimize the leveling-down effect.

Overall, we observe that while GroupDRO and ReSample are effective at improving performance for the worst-performing groups, they tend to degrade performance for the best-performing groups. BM-CL, by leveraging bias mitigation with a continual learning strategy, consistently achieves the smallest degradation in best-group accuracy relative to ERM, while still providing competitive and sometimes superior results for the worst-performing groups. In contrast to EWC, which requires estimating the Fisher information matrix, LwF is computationally more efficient, as it only stores the outputs of the previous stage. We believe that this ability to mitigate the leveling-down effect is crucial for developing robust models that are not only accurate on average, but also fair and consistent across all groups.

\begin{figure*}[ht!]
    \centering
    \includegraphics[width=0.7\linewidth]{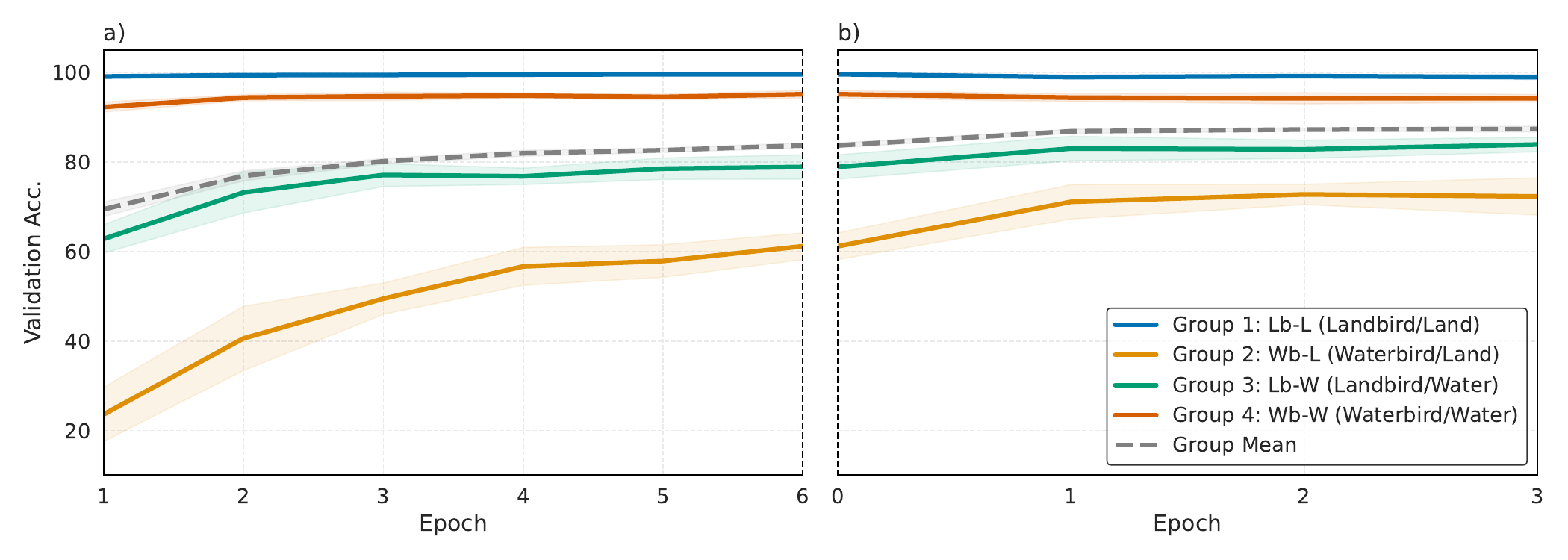}
    \caption{Accuracy (mean $\pm$ standard deviation) in the validation set during training with BM-CL on the Waterbirds dataset. a) Initial training by standard ERM, b) Fine-tuning using BM-CL.}
    \label{fig:waterbirds_training}
\end{figure*}

\begin{table}[t!]
\centering
\caption{Subgroup-level accuracy (mean $\pm$ standard deviation) on the Waterbirds dataset. Subgroups are defined by bird type and background: landbird on land (Lb-L), waterbird on land (Wb-L), landbird on water (Lb-W) and waterbird on water (Wb-W).}
\resizebox{\columnwidth}{!}{
\begin{tabular}{llcccc}
\toprule
\multicolumn{2}{c}{\bf Method} & {\bf Lb-L} & {\bf Wb-L} & {\bf Lb-W} & {\bf Wb-W} \\
\midrule
Baseline & ERM & 99.5 $\pm$ 0.1 & 72.8 $\pm$ 1.3 & 79.6 $\pm$ 1.0 & 94.5 $\pm$ 0.3 \\
\midrule
& GroupDRO & 98.6 $\pm$ 0.2 & 82.6 $\pm$ 0.4 & 86.3 $\pm$ 0.8 & 93.2 $\pm$ 0.5 \\
BM & ReSample & 94.9 $\pm$ 1.1 & 85.5 $\pm$ 1.5 & 87.3 $\pm$ 1.9 & 91.1 $\pm$ 0.4 \\
& JTT & 96.2 $\pm$ 0.5 & 83.5 $\pm$ 1.0 & 81.6 $\pm$ 1.6 & 93.3 $\pm$ 0.5 \\
\midrule
& GroupDRO-LwF & 99.0 $\pm$ 0.2 & 81.6 $\pm$ 1.0 & 82.2 $\pm$ 1.4 & 94.5 $\pm$ 0.4 \\
BM-CL & GroupDRO-EWC & 99.0 $\pm$ 0.3 & 81.2 $\pm$ 1.2 & 82.6 $\pm$ 1.3 & 94.6 $\pm$ 0.5 \\
(ours) & ReSample-LwF & 99.3 $\pm$ 0.2 & 79.5 $\pm$ 1.5 & 81.4 $\pm$ 1.7 & 94.9 $\pm$ 0.5 \\
& ReSample-EWC & 99.3 $\pm$ 0.2 & 78.5 $\pm$ 1.4 & 85.3 $\pm$ 1.5 & 93.8 $\pm$ 0.7 \\
\bottomrule
\end{tabular}}
\label{tab:sota_waterbirds_groups}
\end{table}

\begin{table}[t]
\centering
\caption{Subgroup-level accuracy (mean $\pm$ standard deviation) on the CelebA dataset. Subgroups are defined by hair color and gender: Not blond female (Nh-F), blond female (Bh-F), not blond male (Nh-M), and blond male (Bh-M).}
\resizebox{\columnwidth}{!}{
\begin{tabular}{llcccc}
\toprule
\multicolumn{2}{c}{\bf Method} & {\bf Nh-F} & {\bf Bh-F} & {\bf Nh-M} & {\bf Bh-M} \\
\midrule
Baseline & ERM & 95.7 $\pm$ 0.3 & 87.2 $\pm$ 0.9 & 99.3 $\pm$ 0.1 & 46.1 $\pm$ 2.2 \\
\midrule
& GroupDRO & 92.0 $\pm$ 0.7 & 93.7 $\pm$ 0.5 & 95.8 $\pm$ 0.3 & 72.1 $\pm$ 3.3 \\
BM & ReSample & 90.9 $\pm$ 0.5 & 93.7 $\pm$ 0.9 & 92.9 $\pm$ 0.4 & 80.1 $\pm$ 2.4 \\
& JTT & 86.4 $\pm$ 0.5 & 93.4 $\pm$ 1.8 & 90.8 $\pm$ 1.1 & 76.7 $\pm$ 0.7 \\
\midrule
& GroupDRO-LwF & 92.3 $\pm$ 0.8 & 93.7 $\pm$ 1.1 & 96.5 $\pm$ 0.5 & 72.4 $\pm$ 2.4 \\
BM-CL & GroupDRO-EWC & 92.5 $\pm$ 0.5 & 93.7 $\pm$ 0.6 & 95.9 $\pm$ 0.4 & 75.2 $\pm$ 1.6 \\
(ours) & ReSample-LwF & 91.0 $\pm$ 0.7 & 94.2 $\pm$ 0.9 & 94.3 $\pm$ 0.5 & 80.8 $\pm$ 1.2 \\
& ReSample-EWC & 90.4 $\pm$ 0.6 & 94.8 $\pm$ 0.4 & 93.6 $\pm$ 0.3 & 82.2 $\pm$ 1.3 \\
\bottomrule
\end{tabular}}
\label{tab:sota_celeba_groups}
\end{table}

\begin{table*}[h]
\centering
\caption{Subgroup-level accuracy (mean $\pm$ standard deviation) on the CheXpert dataset. Subgroups are defined by age group (Young, Middle, Old) and presence of pleural effusion (Pe) or absence (Ne).}
\resizebox{0.6\textwidth}{!}{
\begin{tabular}{llcccccc}
\toprule
\multicolumn{2}{c}{\bf Method} & {\bf Ne-Y} & {\bf Pe-Y} & {\bf Ne-M} & {\bf Pe-M} & {\bf Ne-O} & {\bf Pe-O} \\
\midrule
Baseline & ERM & 84.1 $\pm$ 1.6 & 67.5 $\pm$ 2.4 & 78.2 $\pm$ 2.4 & 75.2 $\pm$ 2.3 & 70.9 $\pm$ 2.4 & 77.3 $\pm$ 2.4 \\
\midrule
& GroupDRO & 81.3 $\pm$ 0.9 & 73.0 $\pm$ 1.4 & 77.5 $\pm$ 0.6 & 76.7 $\pm$ 1.4 & 74.0 $\pm$ 0.9 & 75.6 $\pm$ 1.7 \\
BM & ReSample & 81.0 $\pm$ 1.0 & 73.6 $\pm$ 1.3 & 77.3 $\pm$ 1.5 & 75.9 $\pm$ 1.4 & 73.5 $\pm$ 1.7 & 75.7 $\pm$ 1.8 \\
& JTT & 79.0 $\pm$ 1.4 & 67.2 $\pm$ 1.2 & 73.6 $\pm$ 1.6 & 71.1 $\pm$ 1.6 & 67.5 $\pm$ 1.8 & 71.4 $\pm$ 1.7 \\
\midrule
& GroupDRO-LwF & 83.6 $\pm$ 0.9 & 72.4 $\pm$ 1.7 & 78.4 $\pm$ 0.4 & 77.5 $\pm$ 0.3 & 73.7 $\pm$ 1.4 & 77.2 $\pm$ 1.2 \\
BM-CL & GroupDRO-EWC & 81.8 $\pm$ 0.7 & 71.6 $\pm$ 0.6 & 77.6 $\pm$ 1.3 & 75.9 $\pm$ 0.9 & 73.5 $\pm$ 1.4 & 75.7 $\pm$ 1.8 \\
(ours) & ReSample-LwF & 82.8 $\pm$ 1.6 & 73.8 $\pm$ 2.1 & 79.0 $\pm$ 1.1 & 77.0 $\pm$ 1.4 & 74.3 $\pm$ 0.8 & 76.5 $\pm$ 1.2 \\
& ReSample-EWC & 81.5 $\pm$ 1.4 & 72.0 $\pm$ 2.1 & 76.6 $\pm$ 0.5 & 77.0 $\pm$ 1.1 & 71.3 $\pm$ 1.0 & 77.8 $\pm$ 1.2 \\
\bottomrule
\end{tabular}}
\label{tab:sota_chexpert_groups}
\end{table*}

\subsection{Ablation study: impact of pretaining ratio and CL regularization}

We begin our analysis by investigating the contribution of the two main hyperparameters in BM-CL: the pretraining ratio ($\rho$) and the continual learning (CL) regularization strength ($\lambda$). Specifically, we evaluate three values for $\rho$, $0.1$, $0.2$, and $0.3$; and four values for $\lambda$, increasing strengths from no regularization $0.0$, $0.1$, $1.0$, $10.0$. In the case of EWC, since it penalizes deviations in weights directly, it requires a higher regularization value to have an effect. To ensure consistency in the hyperparameter grid, we scale $\lambda$ by a factor of $10^3$ in the case of EWC.

\begin{figure}
    \centering
    \includegraphics[width=\linewidth]{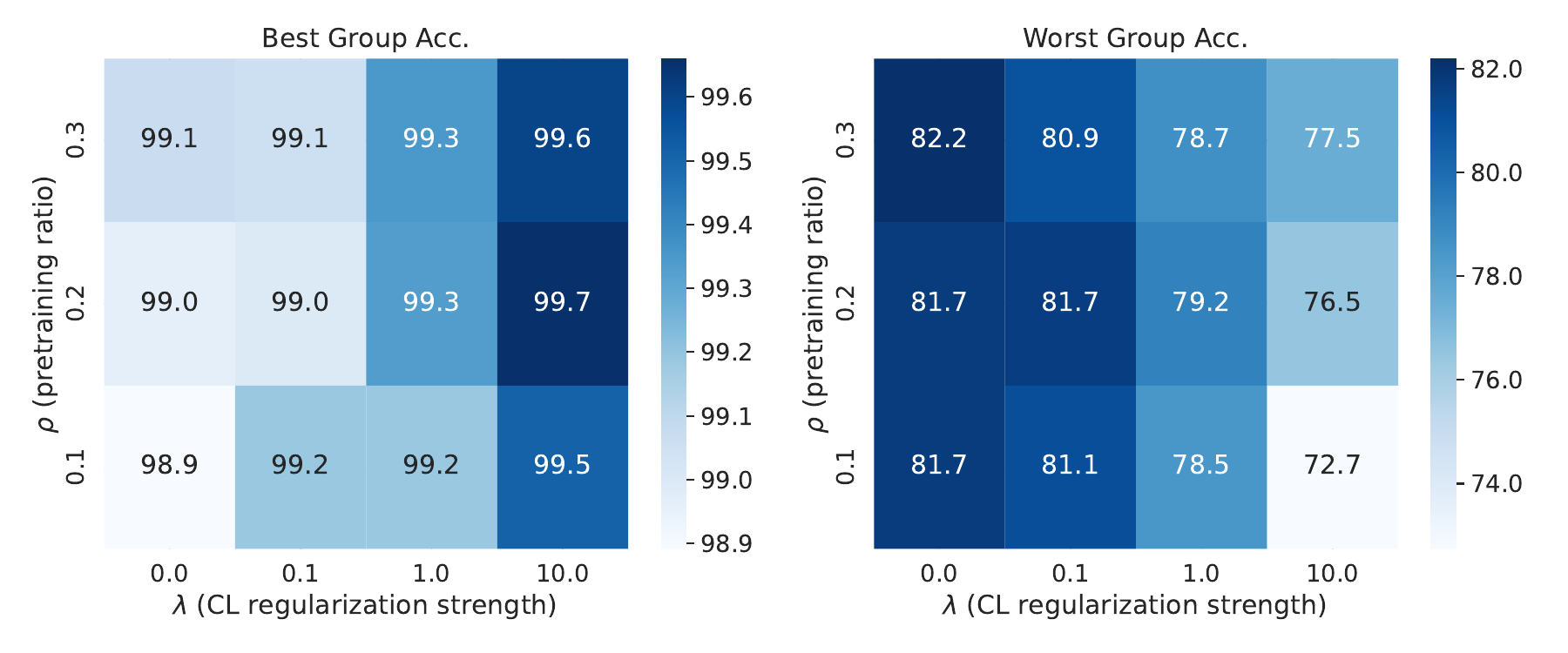}
    \caption{Mean accuracy of BM-CL in the validation set on the Waterbirds dataset comparing the pretraining ratio ($\rho$) and CL regularization strength ($\lambda$) across the best- and the wort-performing groups.}
    \label{fig:waterbirds_ablation}
\end{figure}

Fig. \ref{fig:waterbirds_ablation} presents the mean validation accuracy of BM-CL on the Waterbirds dataset, averaged over 3 independent runs. The evaluation is conducted on both the best- and worst-performing groups, using GroupDRO and LwF as the bias mitigation and continual learning methods, respectively, for the second training stage. Our results reveal the trade-off regarding CL regularization: increasing $\lambda$ helps maintain high performance for the best-performing groups, mitigating the degradation that typically follows in stage 2. In contrast, lower values of $\lambda$ tend to benefit the worst-performing groups, likely because weaker regularization allows the model more flexibility to adapt during bias mitigation. 

We also find no clear trend associated with varying $\rho$, suggesting that its overall impact on performance is limited. An exception occurs for the worst-performing group when $\lambda=10$, where accuracy improves from $72.7$ to $77.5$ as $\rho$ increases from $0.1$ to $0.3$. This behaviour suggests that when regularization is too strong, most of the learning effort shifts to pre-training, making the choice of $\rho$ more influential in such cases.

To better understand how BM-CL supports bias mitigation while reducing catastrophic forgetting, we track validation accuracy during training on the Waterbirds dataset over 5 runs. As illustrated in Fig. \ref{fig:waterbirds_training}, stage 1 consists of standard ERM training. During this phase, the model quickly fits to groups where the target label correlates with the background (i.e., landbird on land and waterbird on water, abbreviated as Lb-L and Wb-W), achieving high accuracy. In contrast, performance on the mismatched groups (waterbird on land and landbird on water: Wb-L and Lb-W) remains lower, which is clear evidence of bias toward these groups. After stage 1, balanced accuracy partitions groups into two disjoint subsets: those where the target class aligns with the background (Lb-L and Wb-W, the best-performing groups), and those where this association does not hold (Wb-L and Lb-W, the worst-performing groups). In stage 2, we fine-tune the model using BM-CL. Here, the CL regularization helps to retain performance on the best-performing groups by preserving their outcomes, while the bias mitigation method (GroupDRO, in this case) guides the model to improve performance on the worst-performing groups. This manifests as increasing validation accuracy on the worst-performing groups during the course of fine-tuning.

\section{Conclusion}

This study introduced Bias Mitigation through Continual Learning (BM-CL), a novel framework that integrates continual learning with bias mitigation strategies to address the prevalent leveling-down effect in machine learning fairness interventions. Our key insight was to reinterpret bias mitigation as a form of task-incremental continual learning, allowing models to improve outcomes for disadvantaged groups while preserving performance for advantaged groups. By incorporating Learning without Forgetting (LwF) and Elastic Weight Consolidation (EWC), BM-CL mitigates the risk of catastrophic forgetting that occurs when fairness objectives shift model priorities.

Our experiments on both synthetic and real-world datasets, including Waterbirds, CelebA and CheXpert, demonstrated that BM-CL consistently improves worst group accuracy while minimizing the performance trade-offs for best group accuracy typically observed in conventional bias mitigation techniques. Notably, LwF-augmented methods preserved best-group accuracy to a greater extent than other approaches, effectively balancing fairness and overall accuracy. Our findings suggest that continual learning principles provide a promising mechanism for developing fairer machine learning models without exacerbating accuracy degradation in advantaged groups. By framing bias mitigation as a continual learning problem, this study opens new pathways for leveraging the extensive toolkit of continual learning techniques to improve fairness without compromising model reliability.

\section{Acknowledgments}
This work was supported by the National Scientific and Technical Research Council (CONICET, Argentina). EF lab was supported by Agencia Nacional de Promoción de la Investigación, el Desarrollo Tecnológico y la Innovación, the Google Award for Inclusion Research and a Googler Initiated Grant. We thank NVidia for the computing resources.

\bibliographystyle{ieeetr}
\bibliography{references}

\end{document}